\documentclass[sigconf]{acmart}
\AtBeginDocument{%
  }

\copyrightyear{2025}
\acmYear{2025}
\setcopyright{acmlicensed}\acmConference[]{}
\acmBooktitle{}
\acmDOI{}
\acmISBN{}

\usepackage{booktabs}
\usepackage{tabularx}
\usepackage{pifont}    
\usepackage{xcolor}    
\usepackage{array}     
\usepackage{balance}
\usepackage{kotex}
\usepackage{hyperref}
\hypersetup{
    colorlinks=true,
    linkcolor=blue,
    filecolor=magenta,      
    urlcolor=cyan,
}

\begin{document}

\title{Roll Your Eyes: Gaze Redirection via Explicit 3D Eyeball Rotation}

\author{YoungChan Choi}
\authornote{These authors contributed equally to this work.}
\email{youngchoy@dankook.ac.kr}
\affiliation{%
  \institution{Dankook University}
  \city{Yongin-si}
  \state{Gyeonggi-do}
  \country{Republic of Korea}
}

\author{HengFei Wang}
\authornotemark[1]
\email{hengfei_wang@163.com}
\affiliation{%
  \institution{University of Birmingham}
  \city{Birmingham}
  \country{United Kingdom}}

\author{YiHua Cheng}
\email{y.cheng.2@bham.ac.uk}
\affiliation{%
  \institution{University of Birmingham}
  \city{Birmingham}
  \country{United Kingdom}}

\author{Boeun Kim}
\authornote{Co-corresponding authors.}
\email{boeun.kim@dankook.ac.kr}
\affiliation{%
  \institution{Dankook University}
  \city{Yongin-si}
  \state{Gyeonggi-do}
  \country{Republic of Korea}}

\author{Hyung Jin Chang}
\email{h.j.chang@bham.ac.uk}
\affiliation{%
  \institution{University of Birmingham}
  \city{Birmingham}
  \country{United Kingdom}}

\author{YoungGeun Choi}
\email{younggch@dankook.ac.kr}
\affiliation{%
  \institution{Dankook University}
  \city{Yongin-si}
  \state{Gyeonggi-do}
  \country{Republic of Korea}}

\author{Sang-Il Choi}
\authornotemark[2]
\email{choisi@dankook.ac.kr}
\orcid{1234-5678-9012}
\affiliation{%
  \institution{Dankook University}
  \city{Yongin-si}
  \state{Gyeonggi-do}
  \country{Republic of Korea}
}

\renewcommand{\shortauthors}{Choi and Wang et al.}

\begin{abstract}
 We propose a novel 3D gaze redirection framework that leverages an explicit 3D eyeball structure. Existing gaze redirection methods are typically based on neural radiance fields, which employ implicit neural representations via volume rendering. Unlike these NeRF-based approaches, where the rotation and translation of 3D representations are not explicitly modeled, we introduce a dedicated 3D eyeball structure to represent the eyeballs with 3D Gaussian Splatting (3DGS). Our method generates photorealistic images that faithfully reproduce the desired gaze direction by explicitly rotating and translating the 3D eyeball structure. In addition, we propose an adaptive deformation module that enables the replication of subtle muscle movements around the eyes. 
 Through experiments conducted on the ETH-XGaze dataset, we demonstrate that our framework is capable of generating diverse novel gaze images, achieving superior image quality and gaze estimation accuracy compared to previous state-of-the-art methods. Codes are available at \url{https://rollyoureyes.github.io/RollyourEyes_ProjectPage}
\end{abstract}


\begin{teaserfigure}
  \includegraphics[width=\textwidth]{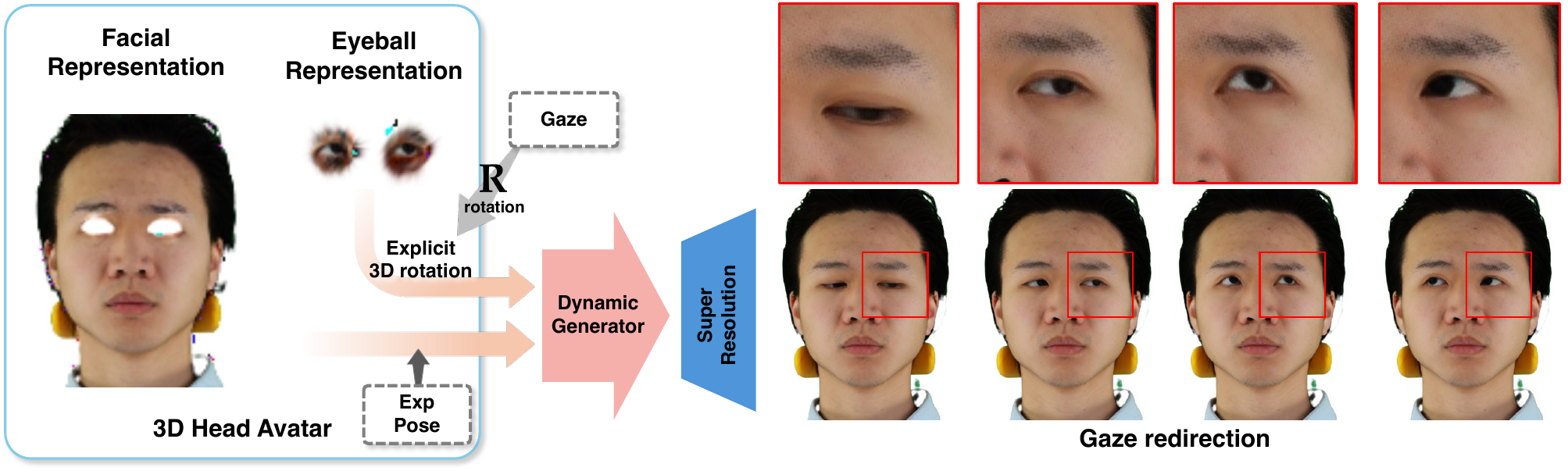}
  \caption{The proposed 3D gaze redirection framework that incorporates an explicit 3D eyeball structure. By rotating an explicit 3D eyeball structure within facial Gaussians, our framework enables photorealistic gaze-redirected image generation.}
  \Description{Facial Representation is modified by expression and pose and Eyeball representation is rotated by gaze vector. Both of the modified representation is fed to the Dynamic Generator, followed by Super Resolution module and generate high-fidelity images}
  \label{fig:teaser}
\end{teaserfigure}

\maketitle

\section{Introduction}
\label{sec:intro}

The human gaze is a critical nonverbal communication channel that goes beyond conveying visual information, playing a significant role in expressing emotions and intentions. In computer vision and graphics, gaze tracking and reproduction technologies are fundamental for various applications, including virtual reality (VR) \cite{piumsomboon2017exploring, david2021towards, pastel2021comparison}, augmented reality (AR) \cite{park2008wearable, bektacs2024gaze, chan2025improving}, remote conferencing \cite{kumar2024enhancing, liang2024emotional, mello2024navigating}, and digital avatars \cite{wang2023high, ruzzi2022gazenerf, li2022eyenerf}. In particular, realistic gaze representation in digital avatars is essential for providing an immersive user experience. 
Recent studies have shown that realistic gaze representation in virtual environments significantly enhances the user experience, particularly by increasing trust and immersion in remote collaboration settings \cite{he2021you}. Furthermore, in virtual social environments such as the metaverse, avatar gaze expression is a crucial element of nonverbal communication, playing a pivotal role in shaping the quality of interactions between users \cite{hennig2023social}.

Gaze redirection is the task of modifying the eye gaze of a subject in an image to generate a photorealistic output that reflects a specified target direction. 
The eyeball has a three-dimensional, sphere-like structure and various gaze directions are represented by physically rotating this sphere-like structure. To achieve realistic eye movements, it is essential for the eyeball representation to behave similarly to actual ocular rotation in human eyes.

However, existing gaze redirection approaches, such as ST-ED \cite{zheng2020self} and GazeNeRF \cite{ruzzi2022gazenerf}, perform gaze manipulation by multiplying implicit eye features with rotation matrices. 
These operations, conducted in the feature space, fail to fully exploit the underlying 3D structure of the eyeball, thereby constraining the accuracy of gaze redirection.
In addition, the implicit neural representation based on volume rendering in neural radiance fields (NeRF) \cite{mildenhall2021nerf} makes explicit rotation and translation of the eyeball infeasible.
In contrast, 3D Gaussian Splatting (3DGS) \cite{kerbl20233d} provides an explicit 3D representation via learned gaussian units which are suitable to represent the 3D structure of eyeball. 

In this paper, we present a novel gaze redirection framework combining 3DGS with an explicit 3D eyeball structure. 
Specifically, we introduce an eye-offset MLP to align the eyeball mesh with expression and pose changes, ensuring anatomically plausible placement.
We further introduce a gaze-guided deformation field to accurately model the subtle movements of the surrounding ocular muscles.
By explicitly rotating the eyeball structure, our method enables flexible and precise gaze redirection, including independent control of the left and right eyes.
In addition to the aforementioned advantages, our method is built upon 3DGS, allowing higher detail preservation and faster rendering speed compared to NeRF-based approaches. 
The main contributions of our method are as follows:
\begin{itemize}
\item We propose a novel 3D gaze redirection framework that leverages explicit 3D eyeball structure. By integrating 3D eyeball structure into a 3DGS-based avatar representation, our model is able to generate realistic gaze-redirected images with high redirection accuracy.      
\item We introduce an gaze-guided adaptive deformation model that dynamically simulates subtle muscle movements around the eyes in response to gaze changes, enhancing the realism of face generation.
\item Our method enables independent gaze control by separately modeling the two eyeballs, expanding the potential of gaze synthesis to include unconventional gaze patterns that are difficult to capture in real-world scenarios.
\end{itemize}
\begin{figure*}[t]
\label{fig:framework}
  \centering
  \includegraphics[width=\linewidth]{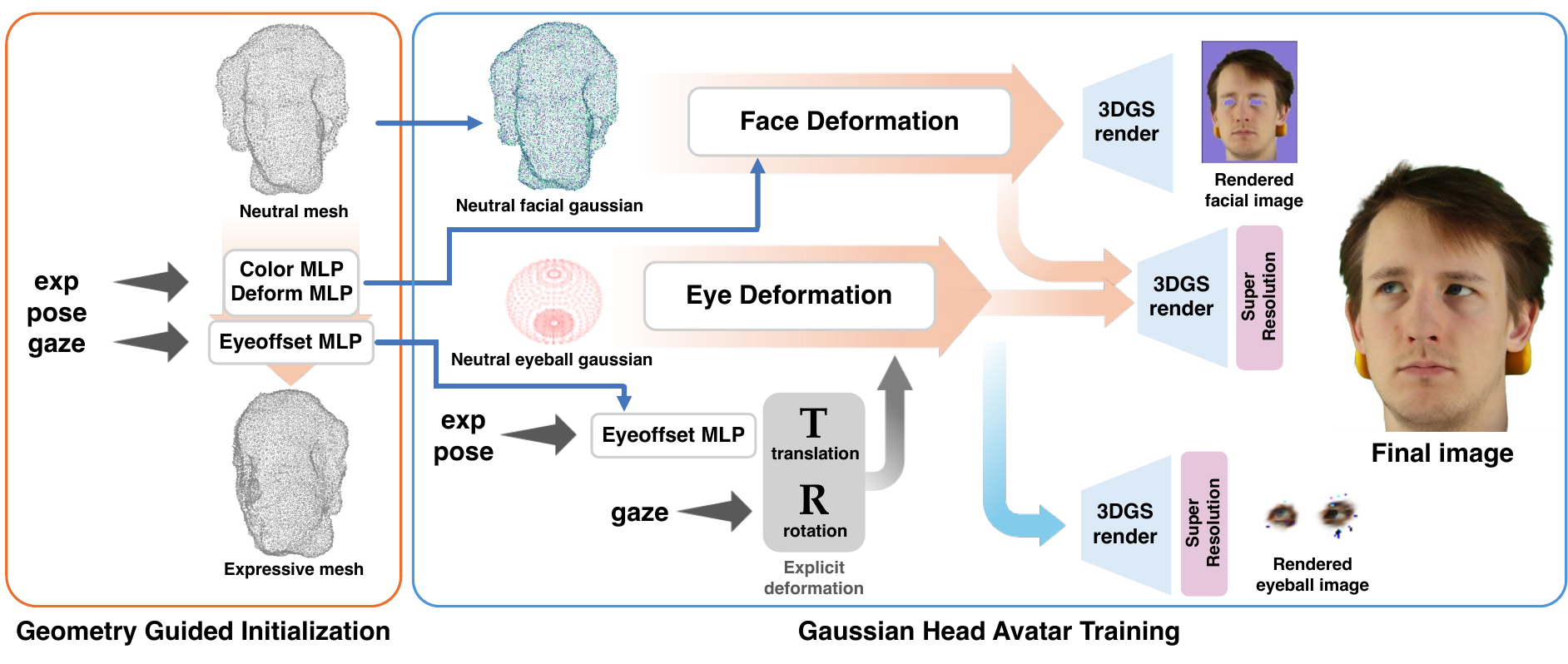}
  \caption{Our framework comprises Geometry-Guided Initialization and Gaussian Head Avatar training. During Gaussian Head Avatar training, we split the process into a face stream for facial deformation and an eyeball stream for explicit eye rotation. Final images are rendered by combining facial and eyeball features via 3D Gaussian Splatting and super-resolution.}
  \label{fig:framework}
\end{figure*}

\section{Related Works}

\subsection{Gaze Redirection}
Gaze redirection involves manipulating the gaze direction in a target image to generate images with a desired gaze. Achieving photorealistic results in gaze redirection requires not only accurately rotating the eyeballs but also capturing the corresponding movements of the eyelids and periocular muscles. The unique challenges of this task arise from the distinct characteristics of the human eye region, where significant information is occluded during movement, and the independent rotation of the eyeballs relative to other facial components. These intrinsic properties necessitate specialized approaches to generate photorealistic images.

Early research on gaze redirection primarily utilized image warping methods. DeepWarp~\cite{ganin2016deepwarp} introduced an efficient warping-based approach that redirects gaze in facial images, enabling real-time performance by learning a warping field from a reference image to a target gaze direction using a deep neural network. ST-ED~\cite{zheng2020self} proposed a spatial transformer-encoder-decoder framework that independently separates and models the eye shape and appearance, allowing for more realistic gaze redirection while preserving subject-specific eye details. ReDirTrans~\cite{jin2023redirtrans} built upon the Spatial-Temporal Encoder-Decoder (ST-ED) framework by introducing a latent-to-latent translation approach, providing more precise and disentangled control over gaze and head redirection. Utilizing a shared latent space enhances appearance and motion consistency while reducing artifacts common in pixel-space transformations.

In parallel, several methods based on Generative Adversarial Networks (GANs) have emerged to improve realism and controllability. He et al.~\cite{he2019photo} introduced a GAN-based gaze redirection framework that employs a gaze-guided warping mechanism within an encoder-decoder generator to synthesize photorealistic eye regions while modeling both eyeball rotation and eyelid deformation. Pan et al.~\cite{chen2021coarse} presented a coarse-to-fine GAN framework guided by both numerical gaze vectors and reference images, utilizing a dual-stage generator and a multi-branch discriminator to enhance both structural accuracy and visual fidelity.

While previous methods primarily operated on 2D image planes, the advent of Neural Radiance Fields (NeRF)~\cite{mildenhall2021nerf} enabled 3D-aware approaches through volumetric modeling. HeadNeRF~\cite{hong2022headnerf} extended NeRF technology to full head modeling with explicit control over facial expressions and gaze, facilitating high-quality novel view synthesis and attribute manipulation through a disentangled representation learning approach. EyeNeRF~\cite{li2022eyenerf} specializes in generating anatomically accurate eye representations with enhanced details for specular highlights and reflections. It also maintains disentangled attribute control while optimizing for realistic eye movement physics and appearance under varying lighting conditions. GazeNeRF~\cite{ruzzi2022gazenerf} addressed the challenge of realistic gaze synthesis by employing a two-stream architecture to separately model the eyeballs and the face, ensuring a more precise and natural integration. DeNeRF~\cite{wang2023high} introduced an eye deformation field that transforms points from the observation space into the canonical space, allowing for more accurate eye movement and deformation modeling.

\subsection{3D Gaussian Splatting}
After the introduction of Neural Radiance Fields (NeRF), reconstructing 3D scenes from multi-view images became significantly feasible. 
However, NeRF-based methods suffer from computational inefficiency, leading to slow training and inference speed. This limitation makes them impractical for real-time or dynamic scene applications.
3D Gaussian Splatting (3DGS)~\cite{kerbl20233d} presents a fundamentally different approach to 3D scene representation. Instead of relying on implicit neural functions, 3DGS models the scene using a set of anisotropic 3D gaussians and achieves real-time rendering through efficient rasterization method. The original 3D Gaussian Splatting framework initializes gaussians from sparse 3D points obtained via Structure-from-Motion (SfM), and optimizes differentiable Gaussian parameters(positions $X$, the multi-channel color $C$, the rotation $Q$, scale $S$ and opacity $A$) to represent the scene. Additionally, for accurate scene reconstruction with finer details, a densification step is performed every 100 iterations to adaptively adds new gaussians during optimization in underrepresented regions. 
Given the camera parameters $\mu$, the gaussians can be rasterized and rendered into a multi-channel image $I$, which can be formulated as below. 
\begin{equation}
    I=R(X,C,Q,S,A;\mu)
\end{equation}

Recently, there has been increasing research on methods for creating animatable avatars using 3D Gaussian splatting~\cite{xu2024gaussian, qian2024gaussianavatars, dhamo2024headgas, shao2024splattingavatar}. These methods can be broadly divided into two main approaches for generating animatable 3D Gaussian avatars.
The first approach involves directly mapping Gaussians onto a 3D mesh model, such as FLAME~\cite{li2017learning}, and dynamically adjusting the Gaussians' position, rotation, and scale according to the mesh's movements. The second approach involves computing a deformation field of the Gaussian attributes, enabling changes in attributes—including color and opacity—based on the avatar's expressions and poses.
The former approach offers a high degree of freedom in generating diverse poses and expressions using the 3D mesh model. However, it is prone to artifacts if a sufficient number of views and a wide range of expression data are not provided. The latter approach generally produces higher-quality images but may occasionally fail to render certain expressions and poses correctly. Additionally, since the deformation field requires extra computation, this method may impact rendering speed.

\section{Method}
\label{sec:Method}

Fig \ref{fig:framework}. shows the overall framework of the proposed method. Our framework consists of two main training stages: Geometry-Guided Initialization and Gaussian Head Avatar Training.
During the initialization stage, we optimize the guidance model following the procedure the original GHA. In this stage, we train the color MLP, deform MLP, and neutral mesh as the guidance model, and additionally the eyeoffset MLP. After Geometry-Guided Initialization, we leverage the neutral mesh to initialize the positions for the Gaussians in the subsequent Gaussian Head Avatar training stage. The learned weights of the color MLP and deform MLP are transferred to initialize the face deformation module, and the weights of the trained eyeoffset MLP are also transferred which is used to determine the positions of the 3D eyeball structure.
In the Gaussian Head Avatar stage, the model is trained to generate the final image. We design the base dynamic generator in GHA to handle face deformation. For accurate and flexible representation of eyeball, we introduce a 3D eyeball structure to form a distinct eyeball stream. We precisely control eye rotation by explicitly rotating the eyeball structure and applying a dedicated dynamic generator to it, which together constitute our eye deformation mechanism.
To generate the final image, the feature maps from both the eye deformation and facial deformation streams are passed through the 3D Gaussian Splatting renderer and the super-resolution module. Additionally, the feature maps from each stream are selectively processed through these modules to produce images used to compute the facial blank loss and eyeball loss.

\subsection{3D Eyeball Structure}

Unlike other facial components, the eyeball is a rotatable structure. To create and animate photorealistic eyeballs, the eyeballs must be represented by a separate 3D structure from the face. 3D mesh models such as FLAME~\cite{li2017learning} attempted to represent the eyeball through a sphere that rotates independently from other facial parts. While this approach was effective for expressing eyeball movements, a sphere is not appropriate to represent the human eyeball, as the eyeball has a slight bulge at the front where the cornea is located. We observed that, in methods such as GaussianAvatars~\cite{qian2024gaussianavatars} and Gaussian Head Avatar~\cite{xu2024gaussian}, the Gaussians representing the eyeball were concentrated in the iris region.

We integrate a precise 3D eyeball structure from 3DGazeNet~\cite{ververas20243dgazenet} into the 3D Gaussian head model, which allocates a higher vertex density to the iris region and fewer vertices to the other areas. In the geometry-guided initialization stage, we positioned the eyeball mesh such that the iris faces forward, aligning it with the 3D eye landmarks of the neutral mesh obtained from an off-the-shelf detector~\cite{bulat2017far}. Next, we introduced eyeball offset MLP(explained in the following sections) to predict eyeball's displacement. The eyeball offset MLP ensures that the eyeball mesh is anatomically repositioned to its appropriate location according to expression changes.

Most gaze datasets provide gaze ground truth as 3D coordinates or 2D coordinates indicating the point-of-focus rather than the actual directional orientation of individual eyeballs. These point-of-focus ground truth reflects both optical and anatomical factors~\cite{tabernero2007mechanism}, including the kappa angle~\cite{lefohn2003ocularist} (the angle between the visual and optical axes) and individual variations in eye structure. If we rotate the eyeball directly using the coordinate-based gaze ground truth provided by the dataset, a misalignment between the rotated eyeball and the actual gaze direction can be observed. This discrepancy arises due to the kappa angle, and such misalignment negatively affects the optimization of the Gaussians.

To ensure minimal discrepancy between the simulated eyeball direction and the actual anatomical orientation, we guide the orientation of our eyeball mesh using a pseudo-ground truth (GT) direction, which is obtained from the directional outputs of 3DGazeNet~\cite{ververas20243dgazenet}.
This pseudo-GT direction offers two key advantages. First, it allows gaze redirection on datasets without gaze ground truth. Second, it is suitable for pixel-based optimization. As it is obtained by computing the rotation between the initially aligned eyes and the 3D-lifted iris center and by rotating the eyeballs accordingly, the eyeball direction derived from 3DGazeNet yields more reasonable results in terms of pixel-level alignment.

\subsection{Avatar Representation}

\textbf{Face Deformation}
We follow GHA~\cite{xu2024gaussian} for 3D Gaussian head representation. We first extract the neutral mesh from DMTet and construct a canonical neutral Gaussians for neutral expression with attributes $\{X_0, F_0, Q_0, S_0, A_0\}$, where $X_0 \in \mathbb{R}^{N \times 3}$ represents positions, $F_0 \in \mathbb{R}^{N \times 128}$ denotes point-wise feature vectors, and $Q_0 \in \mathbb{R}^{N \times 4}$, $S_0 \in \mathbb{R}^{N \times 3}$, and $A_0 \in \mathbb{R}^{N \times 1}$ represent rotation, scale, and opacity respectively.

Then, we implement an MLP-based, expression-conditioned dynamic generator to function as a face deformation field $\Phi_f$ to get the attributes of Gaussians on target expression. 
\begin{equation}
(X,C,Q,S,A)=\Phi_f(X_0, F_0, Q_0, S_0, A_0; \theta, \beta)
\end{equation}
Generally, Gaussians representing 3D scene are represented by their positions $X$, the multi-channel color $C$, the rotation $Q$, scale $S$, and opacity $A$
The position of Gaussians $X$ are calculated by adding displacement predicted by deformation MLPs $D_{\text{exp}}, D_{\text{pose}}$.
\begin{equation}
    \begin{split}
    X = X_0 &+ \lambda_{\text{exp}}(X_0)\, D_{\text{exp}}(X_0, \text{exp}) \\
    &+ \lambda_{\text{pose}}(X_0)\, D_{\text{pose}}(X_0, \text{pose}),
    \end{split}
\end{equation}
\begin{equation}
\label{eq:exp_influence}
    \lambda_{\text{exp}}(x) = 
    \begin{cases}
        1, & \text{dist}(x, P_0) < t_1 \\
        \frac{t_2 - \text{dist}(x, P_0)}{t_2 - t_1}, & \text{dist}(x, P_0) \in [t_1, t_2] \\
        0, & \text{dist}(x, P_0) > t_2
    \end{cases}
\end{equation}
$\lambda_{\text{exp}}(\cdot)$ and $\lambda_{\text{pose}}(\cdot)$ represent the amount of influence from expression and head pose. They are determined based on the minimum distance $\text{dist}(x, P_0)$ between each Gaussian position $x$ and the set of 3D landmarks $P_0$ (excluding the eyes).
If $\text{dist}(x, P_0) < t_1$, the attributes of the Gaussian at $x$ are fully driven by the expression parameters. Conversely, if $\text{dist}(x, P_0) > t_2$, indicating that $x$ is far from all 3D landmarks, its attributes are entirely determined by the pose vector.
We use two color MLPs, $C_{\text{exp}}$ and $C_{\text{pose}}$, to predict the multi-channel color $C$. Similarly, rotation, scale, and opacity are predicted using two separate MLPs: $A_{\text{exp}}$ and $A_{\text{pose}}$.
\begin{equation}
C = \lambda_{\text{exp}}(X_0)\, C_{\text{exp}}(F_0, \theta) + \lambda_{\text{pose}}(X_0) C_{\text{pose}}(F_0, \beta).
\end{equation}
\begin{equation}
\begin{split}
    \{Q, S, A\} = \{Q_0, S_0, A_0\} &+ \lambda_{\text{exp}}(X_0)\, A_{\text{exp}}(F_0, \theta) \\ &+ \lambda_{\text{pose}}(X_0) \,A_{\text{pose}}(F_0, \beta).
\end{split}
\end{equation}
After transforming the Gaussians from the neutral to the target expression using a series of MLPs $\Phi_f(\cdot)$ parameterized by $\theta$ and $\beta$, we rasterize them using $R(\cdot)$ to generate the facial feature map $M_f$:
\begin{align}
M_f &= R(\{X, C, Q, S, A\}) \\ 
&= R(\Phi_f(X_0, F_0, Q_0, S_0, A_0; \theta, \beta)),
\end{align}

\textbf{Eye Deformation}
We integrate the 3D eyeball mesh into a neutral mesh by aligning the 3D eyeball structure with the center of the 3D eye landmarks and orienting it to face forward. Then, following the method used in GHA, we initialized Gaussians on the eyeball region within this neutral mesh.
We observed that the dynamic generator of GHA fails to learn the correct eyeball positions. To ensure that the Gaussians representing the eyeballs are located at the anatomically correct regions within the mesh, we introduce an eyeoffset MLP, $\mathcal{D}_e$, that predicts the displacement of the eyeball center, $\Delta P_{eyeball}$, based on expression and pose. The adjusted eyball center, $X_{eyeball}$ is written as:
\begin{align}
X_{eyeball} &= X_{eyeball}^0 + \Delta P_{eyeball}, \\
\Delta P_{eyeball} &= \mathcal{D}_e(exp, pose),
\end{align}
where $exp$ and $pose$ denote expression vector and pose vector, respectively.
The positions of the Gaussians representing the eyeballs are then uniformly adjusted based on the predicted displacement.
The eye-offset MLP is trained throughout the framework, beginning with anatomical displacement prediction during geometry-guided initialization and further refined during avatar training to support precise and photorealistic gaze synthesis.

We observed that the Gaussian representation of the eyeball tends to lose its inherent spherical structure during the training of the Gaussian Head Avatar. Since accurate gaze redirection requires the eyeball Gaussians to preserve their original shape and maintain semantic consistency across varying expressions and gaze directions, structural stability is crucial. To ensure this, the eyeball Gaussians are positioned based on the combined effects of anatomical displacement from the eye-offset MLP and gaze-driven rotation. During Gaussian Head Avatar training, we further optimize the color, rotation, scale, and opacity of the eyeball Gaussians to enhance photorealism in the final rendering.

\subsection{Gaze Guided Deformation Field}
\label{sec:gaze guided deformation field}
When changing the direction of gaze, humans subtly move the muscles around their eyes~\cite{porter1995extraocular}. However, most of the existing methods~\cite{jin2023redirtrans, hong2022headnerf, ruzzi2022gazenerf, li2022eyenerf, wang2023high} that utilize explicit eyeball structures focus exclusively on rotating the eye structure itself, neglecting the associated movements of the surrounding muscles. To generate photorealistic images, capturing these subtle movements of the periocular muscles is essential.
Inspired by $\lambda_{exp}(\cdot)$ and $\lambda_{pose}(\cdot)$ in GHA~\cite{xu2024gaussian}, we introduce an additional influence factor, $\lambda_{gaze}(\cdot)$. This factor determines the extent to which each Gaussian is influenced by the gaze vector, with the influence decreasing as the minimum distance between the Gaussian position $x$ and the 3D eye landmarks $P_{\text{eye}}$ increases. Gaussians closer to the eye landmarks are thus more strongly modulated by gaze direction.
Specifically, we define the gaze influence function $\lambda_{gaze}$ as below:
\begin{equation}
\label{eq:gaze_influence}
\begin{split}
\lambda_{\text{gaze}}(x) = 
    \begin{cases}
        1, & \text{dist}(x, P_{\text{eye}}) < t_3 \\
        \frac{t_4 - \text{dist}(x, P_{\text{eye}})}{t_4 - t_3}, & \text{dist}(x, P_{eye}) \in [t_3, t_4] \\
        0, & \text{dist}(x, P_{\text{eye}}) > t_4 
    \end{cases}
\end{split}
\end{equation}
Similar to $t_1$ and $t_2$, $t_3$ and $t_4$ define the minimum and maximum distances for gaze influence. If $\text{dist}(x, P_{\text{eye}}) < t_3$, the Gaussian at position $x$ is fully influenced by the gaze vector, whereas values greater than $t_4$ indicate no gaze influence.
By incorporating $\lambda_{gaze}(\cdot)$ to our framework, we are able to depict not only eye rotation but also the deformations of the eyelids and periocular skin that naturally accompany changes in gaze direction. This comprehensive approach significantly enhances the realism of synthesized facial images, particularly in the eye region, which plays a critical role in human perception of authenticity in facial representations.

\begin{table*}
  \caption{Quantitative Comparisons. We compared our model with state-of-the-art methods using standard image quality metrics (SSIM, PSNR, LPIPS, FID), and observed consistent improvements across all metrics. The head area denotes scores over the entire head, while the face area focuses on the region excluding hair and neck.}
  \label{tab:commands}
  \begin{tabularx}{\textwidth}{
  >{\centering\arraybackslash}p{2.8cm}|
  >{\centering\arraybackslash}p{1.5cm}
  >{\centering\arraybackslash}p{1.5cm}
  >{\centering\arraybackslash}p{1.5cm}
  >{\centering\arraybackslash}p{1.5cm}
  >{\centering\arraybackslash}p{1.5cm}
  >{\centering\arraybackslash}p{1.5cm}
  >{\centering\arraybackslash}p{2.7cm}}
    \toprule
    Method & 3D-based & SSIM $\uparrow$ & PSNR $\uparrow$ & LPIPS $\downarrow$ & FID $\downarrow$ & Gaze $\downarrow$ & Identity Similarity $\uparrow$\\
    \midrule
    ST-ED \cite{zheng2020self} & \textcolor{red}{\ding{55}} & $0.726$ & $17.530$ & $0.300$ & $115.020$ & $16.217$ & $24.347$\\
    HeadNeRF \cite{hong2022headnerf} & \textcolor{green}{\ding{51}} & $0.720$ & $15.298$ & $0.294$ & $69.487$  & $12.117$ & $46.126$\\
    GazeNeRF \cite{ruzzi2022gazenerf} & \textcolor{green}{\ding{51}} & $0.733$ & $15.453$ & $0.291$ & $81.816$  & $6.944$ & $45.207$\\
    \hline
    Ours (head area) & \textcolor{green}{\ding{51}} & $0.814$ & $22.498$ & $0.117$ & \textbf{31.272}  & $5.696$ & $91.930$\\
    Ours (facial area) & \textcolor{green}{\ding{51}} & \textbf{0.905} & \textbf{30.292} & \textbf{0.062} & \textbf{31.272}  & \textbf{5.006} & \textbf{91.947}\\
    \bottomrule
  \end{tabularx}
\end{table*}

\begin{figure*}[t]
  \centering
  \includegraphics[width=\linewidth]{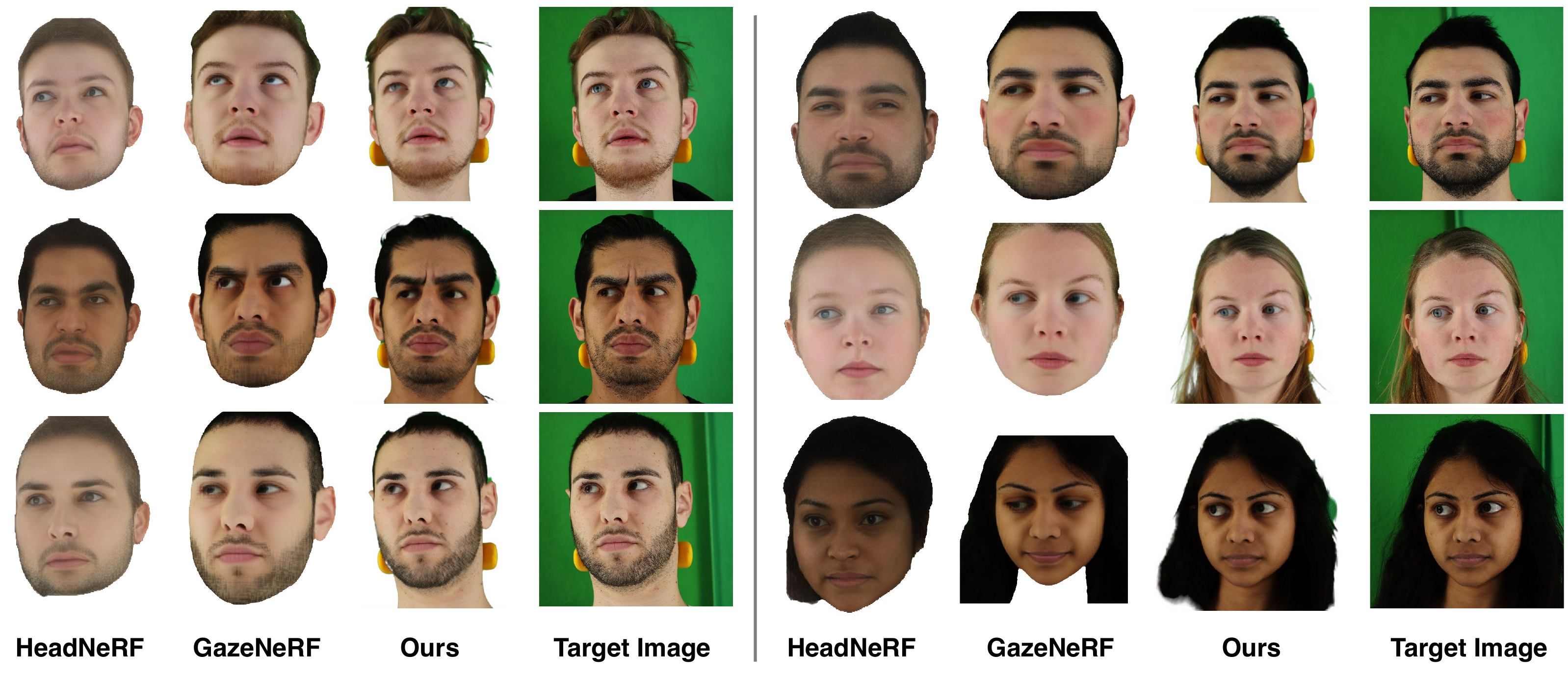}
\caption{Qualitative comparisons on ETH-XGaze dataset\cite{zhang2020eth}. NeRF based methods, HeadNeRF~\cite{hong2022headnerf} and GazeNeRF~\cite{ruzzi2022gazenerf} struggles to reconstruct fine details on the subject's face. Our framework is able to reconstruct more photorealistic images with full of facial details such as mustache and the texture of iris.}
  \Description{Comparison with state-of-the-art}
  \label{fig:comparison}
\end{figure*}

\section{Training}
In this section, we descibe the training pipeline and loss function of our framework. 
The training procedure is divided into two stages: Geometry-Guided Initialization and Gaussian Head Avatar training.

\subsection{Geometry-Guided Initialization}

\begin{figure*}[t]
\label{fig:independent_control}
  \centering
  \includegraphics[width=\linewidth]{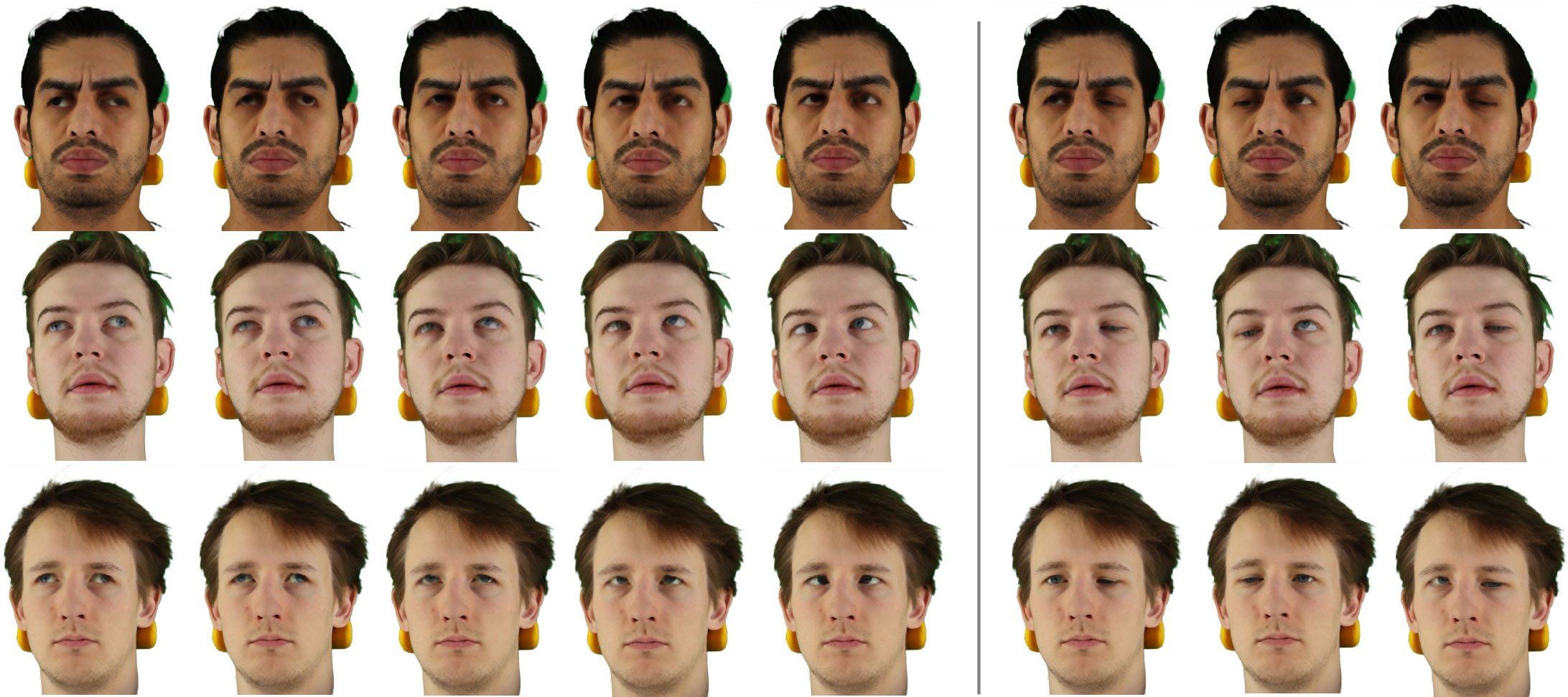}
  \caption{Visualization of independent gaze control. The left image shows gazes redirected from outward to inward directions, while the right shows vertically divergent gazes with one eye looking up and the other down.}
  \label{fig:independent_control}
\end{figure*}

In the initialization stage, we aim to determine the initial 3D coordinates of Gaussians in neutral expression. Following the GHA methodology, we optimized a mesh guidance model to provide reliable initialization for the Gaussians. In GHA, two color MLPs $C_{\text{exp}}$, $C_\text{pose}$ and two deformation MLPs $D_\text{exp}$, $D_\text{pose}$ are trained to predict the per-vertex 32-channel color and displacements of each vertex positions respectively. We integrated gaze guided deformation field $D_\text{leftgaze}$, $D_\text{rightgaze}$, $C_\text{rightgaze}$, $C_\text{rightgaze}$ following \ref{sec:gaze guided deformation field} to express the subtle movements of the periocular muscles. 

We integrated the training of eyeoffset MLP with the training of mesh guidance model.
To train the EyeOffset MLP to predict eyeball displacement caused by expression and pose changes, we use an L1 loss between the predicted and ground truth displacements of 3D eye landmark centers from the neutral and target meshes.
The loss of mesh guidance model consists of L1 image loss $\mathcal{L}_{RGB}$, a mask IoU loss $\mathcal{L}_{sil}$, an L2 loss on 3D facial landmark $\mathcal{L}_{def}$, and three constraints: $\mathcal{L}_{offset}$, $\mathcal{L}_{lmk}$, $\mathcal{L}_{lap}$. $\mathcal{L}_{offset}$. $\mathcal{L}_{offset}$ penalizes non-zero displacements to avoid global offsets, $\mathcal{L}_{lmk}$ enforces SDF values at 3D landmarks to be near zero for surface alignment, and $\mathcal{L}_{lap}$ maintains mesh smoothness via Laplacian regularization.
The final loss of geometry-guided initialization is formulated as:
\begin{equation}
\begin{split}
\mathcal{L} = & \lambda_{RGB}\mathcal{L}_{RGB} + \lambda_{sil}\mathcal{L}_{sil} + \lambda_{def}\mathcal{L}_{def} + \lambda_{offset}\mathcal{L}_{offset} \\ &+ \lambda_{lmk}\mathcal{L}_{lmk} + \lambda_{lap}\mathcal{L}_{lap} + \lambda_{eyeoffset}\mathcal{L}_{eyeoffset}
\end{split}
\end{equation}
$\lambda$ denotes the weights of each term, with $\lambda_{RGB}=0.1$, $\lambda_{sil}=0.1$, $\lambda_{def}=1$, $\lambda_{offset}=0.01$, $\lambda_{lmk}=0.1$, $\lambda_{lap}=100$ and $\lambda_{eyeoffset}=0.1$
Finally, we initialize the Gaussian model using the trained mesh guidance model. The position of the Gaussians are initialized based on the vertex positions in the neutral pose, and a per-vertex feature vector extracted from mesh guidance model is stored for Gaussian Head Avatar Training. We retain the trained MLPs($D_\text{exp}$, $D_\text{pose}$, $C_\text{exp}$, $C_\text{pose}$, $C_\text{leftgaze}$, $C_\text{rightgaze}$, $C_\text{rightgaze}$, $C_\text{rightgaze}$) for Gaussian model.
When training geometry-guided initialization, we calculate the influence factor using equation~\ref{eq:exp_influence} and equation~\ref{eq:gaze_influence} while $t_1=0.15$, $t_2=0.25$, $t_3=t_1\times 0.5$, and $t_4=t_2\times 0.5$.

\subsection{Gaussian Head Avatar Training}
\label{sec:Gaussian Head Avatar Training}
During Gaussian Head Avatar Training, Gaussian model is optimized using three outputs: rendered facial Gaussian image, rendered eyeball Gaussian image and rendered final image. They are generated following the procedure explained in \ref{sec:Method}. We designed the GHA training loss to satisfy two key criteria. First, the Gaussians representing the eyeballs should affect only the pixels corresponding to the eyeballs. Second, the Gaussians representing other facial regions, excluding the eyeballs, should have minimal influence on the eyeball pixels. To satisfy the above criteria, we build our loss upon the GHA's loss function. 

\textbf{Image Synthesis Loss}
In GHA, the loss is defined as a combination of L1 loss and VGG perceptual loss~\cite{zhang2018unreasonable} between the generated images $I_{hr}, I_{lr}$ and the ground truth $I_{gt}$ where $I_{lr}$ denotes the first three channels of rendered image and $I_{hr}$ denotes the final rendered image generated by super-resolution module. The total loss of GHA in Gaussian Head Avatar Training phase is:
\begin{equation}
\begin{split}
\mathcal{L} = ||I_{hr}-I_{gt}||_1 + \lambda_{vgg}VGG(I_hr,I_gt) + \lambda_{lr}||I_{lr}-I_{gt}||_1
\end{split}
\end{equation}

\textbf{Eyeball Loss}
Our framework employs a two-stream architecture that processes eyeball and facial information separately. When training two-stream architecture, it is important for each stream to accurately perform its designated role. Specifically in our framework, the eyeball stream should concentrate exclusively on eyeball-related information, while the facial stream should concentrate on representing facial features only.
To guarantee that, we first introduced two eyeball loss terms based on the eye region mask.
\begin{equation}
    \mathcal{L}_{eyeball} = \lambda_{lr}\mathcal{L}_{eyeLR}+\lambda_{hr}\mathcal{L}_{eyeHR}
\end{equation}
\begin{equation}
    \mathcal{L}_{eyeLR} = ||M_e \odot I_{lr}-M_e \odot D(I_{gt})||_1
\end{equation}
\begin{equation}
    \mathcal{L}_{eyeHR} = ||M_e \odot I_{hr}-M_e \odot I_{gt}||_1
\end{equation}
$\mathcal{L}{eyeLR}$ and $\mathcal{L}{eyeHR}$ denote the low-resolution and high-resolution eyeball losses, respectively. $M_e$ is the eye region mask, $I_{lr}$ is the low-resolution eyeball rendering, $I_{hr}$ is the super-resolved high-resolution output, and $I_{gt}$ is the ground truth image.
$\mathcal{L}_{eyeLR}$ is formulated as the L1 loss between the masked eye region of the feature map, rendered using only the Gaussians from the eyeball stream, and the masked eye region of the downsampled ground truth image $D(I_{gt})$.
The high-resolution eyeball loss is similarly defined as the L1 loss between the masked eye region of the final image generated by the super-resolution module and the masked eye region of the original ground truth image. The feature map is rendered using only Gaussians from the eyeball stream.

\textbf{Facial Blank Loss}
When training with the eyeball loss, we observed that the facial stream interferes with the eye region. To disentangle the eyeball and facial representations, we introduce the facial blank loss, which penalizes the influence of facial Gaussians within the eye area. Specifically, we render the image using only the facial stream Gaussians, denoted as $I_{facial}$, and compute an L1 loss between the eye-masked $I_{facial}$ and a randomly assigned background color $c$ to avoid white artifacts in the eye region:
\begin{equation}
\mathcal{L}_{blank} = ||M_e \odot I_{facial}-M_e \odot c||_1
\end{equation}

\textbf{Final Loss}
The combined eyeball loss aggregates the above loss functions as follows:
\begin{equation}
\mathcal{L}_{eyeball} = \lambda_{lr}\mathcal{L}_{eyeLR}+\lambda_{hr}\mathcal{L}_{eyeHR}+\lambda_{blank}\mathcal{L}_{blank}.
\end{equation}
\begin{equation}
    \mathcal{L} = \mathcal{L}_{RGB} + \mathcal{L}_{eyeball}
    \label{eq:total}
\end{equation}
The weights were set to $\lambda_{lr}=0.5$, $\lambda_{hr}=0.5$, and $\lambda_{blank}=1.0$ empirically.
Finally, we defined the final loss by integrating the combined eyeball loss$\mathcal{L}_{eyeball}$ and image synthesis loss $\mathcal{L}_{RGB}$.
By incorporating these loss functions, we established a model architecture in which the eyeball and facial streams fulfill their roles accurately without interfering with each other.
\section{Experiment}

\newcolumntype{Y}{>{\centering\arraybackslash}X}

We trained our model on a single RTX 3090 GPU with 24GB of VRAM. We trained our model for 500 epochs and the entire training process takes approximately six hours.

\subsection{Dataset}

To demonstrate the effectiveness of our framework, we conducted the experiment on ETH-XGaze dataset. The ETH-XGaze dataset is an extensive and comprehensive resource for gaze research which includes 110 subjects, including images from 18 different camera views. It offers 80 subjects designated for training, with total of 756,000 training frames. We selected 20 subjects from the training dataset for our experiments, using frames with consistent illumination for better optimization. We followed the preprocessing pipeline established by GHA~\cite{xu2024gaussian}. 

\subsection{Comparison with the state-of-the-art}

\textbf{Baselines}
We compare our method against the state-of-the-art gaze redirection model including 2D-based method ST-ED, and 3D-based method HeadNeRF and GazeNeRF.

\textbf{Evaluation Metrics}
To assess the quality of reconstructed images, we used four different metrics, including SSIM (Structural Similarity Index Measure), PSNR (Peak Signal-to-Noise Ratio), LPIPS (Learned Perceptual Image Patch Similarity), and FID (Fréchet Inception Distance). Following the evaluation method on GazeNeRF, we measured gaze angular errors with ResNet50~\cite{he2016deep}-based estimator and identity similarity measured based on the face recognition model from FaceX-Zoo~\cite{wang2021facex}. As noted in the limitations of GHA, rendering failures often occur for subjects with long hair, influenced by the limited expression diversity in the ETH-XGaze dataset. To address this, we calculated both a head area score (entire head) and a facial area score (excluding the hair and neck region). This additional metric provides a more accurate evaluation of gaze generation performance by focusing on the facial region, which is more relevant to the task.

\textbf{Reenactment Capability}
To assess our method's reenactment performance, we conducted experiments using expression parameters, pose parameters, and gaze vectors from frames not included in the training dataset. These parameters were fed into our trained Gaussian head model to generate images for evaluation.
As shown in Table 1, our method significantly outperforms existing state-of-the-art approaches across all evaluation metrics, indicating not only the high-fidelity and photorealism of the generated images, but also accurate gaze redirection and strong identity preservation.

The qualitative results in Fig~\ref{fig:comparison} demonstrate that our method produces images with significantly higher detail and quality than other methods.
Additionally, Our model could render fine details such as facial hair and iris textures specific to the subject. This level of detail contributes to the generation of photorealistic images and significantly improves the visual quality of the generated results.

\begin{figure}[t]
  \centering
  \includegraphics[width=\linewidth]{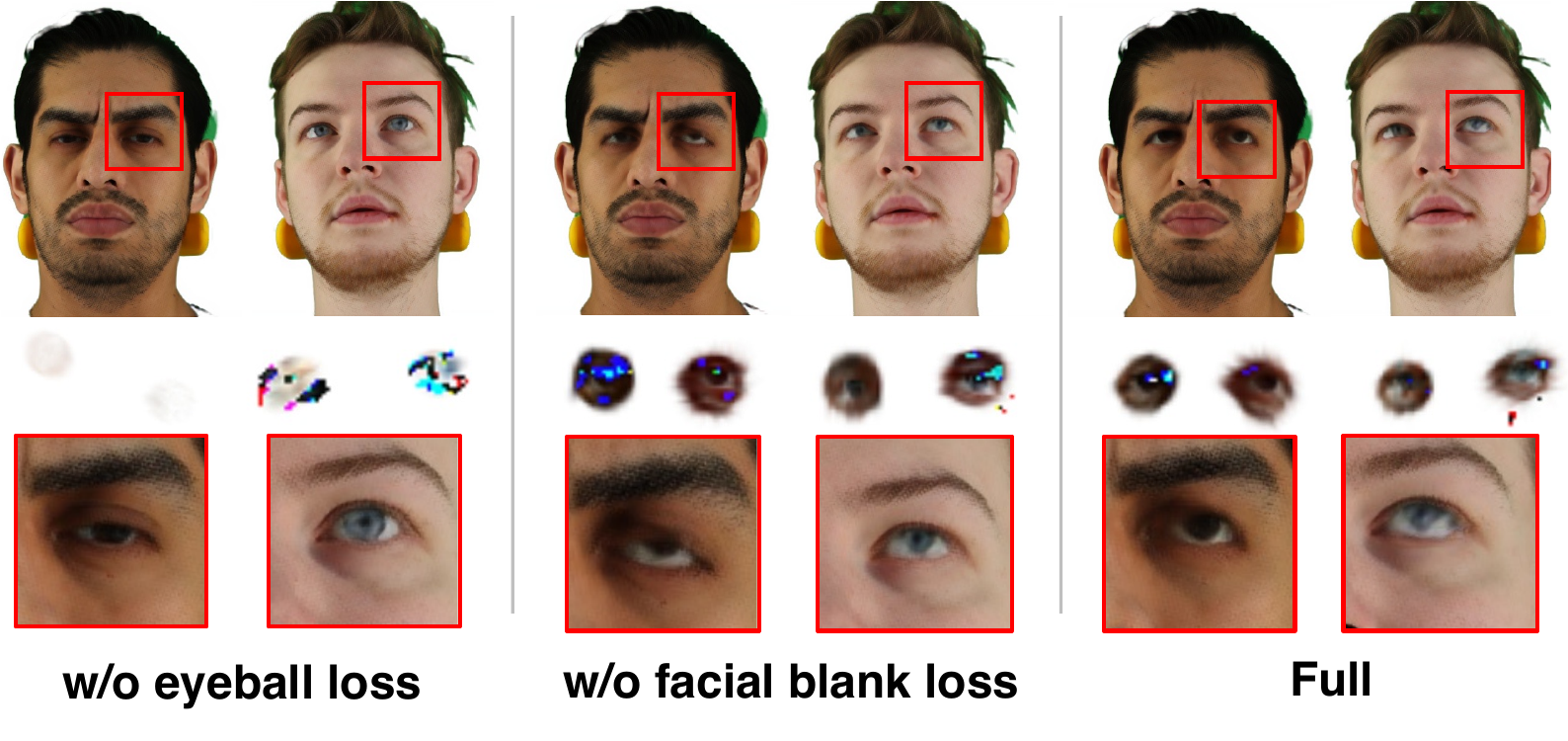}
  \caption{
  Ablation study results on eyeball loss ($\mathcal{L}_{eyeball}$) and facial blank loss ($\mathcal{L}_{blank}$).}
  \label{fig:ablations}
\end{figure}
\begin{table}
  \centering
  \caption{Quantitative results of ablation studies evaluating the impact of the facial blank loss and the eyeball loss}
  \label{tab:abl12}
  \begin{tabularx}{\columnwidth}{>{\centering\arraybackslash}p{4cm}|YYY}
    \toprule
    Method & SSIM$\uparrow$ & PSNR$\uparrow$ &  Gaze$\downarrow$ \\
    \midrule
    w/o facial blank loss & \textbf{0.8305} & 24.114 & 5.398 \\
    w/o eyeball loss  & 0.8284 & 24.149& 4.910 \\
    Full version & \textbf{0.8305} & \textbf{24.166} & \textbf{4.407} \\
    \bottomrule
  \end{tabularx}
\end{table}

\textbf{Novel Gaze Generation}
We utilized expression and pose parameters from a randomly selected frame and generated images with varying gaze vectors. The model accepts two gaze vectors, the left gaze vector and right gaze vector, generating the images with independent eye control. Fig~\ref{fig:independent_control} showcases the results across a broad spectrum of gaze directions and subjects. Our method produces photorealistic images across diverse gaze variations. Furthermore, it accurately captures subtle muscle movements related to gaze direction, enabling natural and consistent changes in the periocular region in various gaze. These fine-grained details significantly enhance the fidelity of the generated images.
Thanks to the explicit 3D eyeball structure and the independent control of each eye, our method enables precise and per-eye gaze manipulation, as demonstrated in Fig~\ref{fig:independent_control}. It can generate photorealistic images even in challenging scenarios where the eyes exhibit cross-eyed or other unconventional gaze directions.
This fine-grained control allows for the synthesis of eye gaze patterns that are difficult or even impossible to capture in real-world settings. By supporting the generation of realistic yet unconventional gaze behaviors, our method expands the creative possibilities of gaze image synthesis and demonstrates strong potential for advancing the field.

\begin{table}
  \caption{Quantitative ablation studies evaluating the impact of the gaze-guided deformation field.}
  \label{tab:abl3}
  \begin{tabularx}{\columnwidth}{>{\centering\arraybackslash}p{4.5cm}|*{3}{>{\centering\arraybackslash}m{0.9cm}}}
    \toprule
    Method & SSIM$\uparrow$ & PSNR$\uparrow$ & FID$\downarrow$ \\
    \midrule
    w/o Gaze-guided deformation field & 0.8089 & 19.1622 & 32.9459 \\
    Full version & \textbf{0.8108} & \textbf{19.3932} & \textbf{31.2309} \\
    \bottomrule
  \end{tabularx}
\end{table}

\subsection{Ablation Studies}

We conducted an ablation study to assess the effectiveness of our newly introduced loss functions and gaze-guided deformation field in the gaze generation framework. We selected 10 subjects in EHT-XGaze dataset and analyzed the impact of each module. Table~\ref{tab:abl12} and Fig~\ref{fig:ablations} presents a comparative analysis of three configurations: (1) without eyeball loss, (2) without facial blank loss, and (3) our complete method incorporating both losses. Without the eyeball loss, we observed that the rendered Gaussians for the eyeball region appeared either nearly transparent or visually distorted, indicating that facial Gaussians were being deformed to compensate for the eye rendering. This often led to inaccurate gaze synthesis, particularly for downward gaze directions. The facial blank loss further mitigates these interference by improving the rendering quality around the eyeball region. As demonstrated in \ref{tab:abl12}, the combination of these loss functions results in improved gaze accuracy and overall image quality.
In \ref{tab:abl3}, we present an ablation study evaluating the effectiveness of the gaze-guided deformation field. 
To assess whether our gaze deformation field accurately captures subtle muscle movements around the eye region, we cropped the eye and surrounding areas from both the ground truth and the rendered images and masked out the eyeball region in white to isolate subtle muscle deformations. The presence of the gaze-guided deformation field led to consistent improvements across SSIM, PSNR, and FID metrics, indicating that it significantly enhances the realism of image generation, particularly in the eye region.
\newcolumntype{Y}{>{\centering\arraybackslash}X}

\section{Conclusion}

In this paper, we propose a novel gaze redirection framework that incorporates explicit 3D eyeball structure into 3DGS. To achieve realistic eye movements, we apply physical rotation and translation to dedicated 3D eyeball representation to similarly model acutal ocular roation in human eyes.
We also introduce adaptive deforamtion field to simulate the subtle muscle movements around the eyes according to directed gaze vector. 
We have shown that our proposed framework outperforms existing gaze redirection methods in terms of image quality and gaze redirection accuracy, while also demonstrating strong identity preservation. Our framework enables the generation of unconventional gaze images that are infeasible to capture in the real world, thanks to its accurate redirection and independent gaze control capabilities. We believe that our work expands the potential of gaze image synthesis by enabling the generation of gaze patterns beyond the limitations of real-world data, and paves the way for future applications in dynamic avatar control, human-computer interaction, and virtual reality systems.

\begin{acks}
This research was supported by the MSIT(Ministry of Science and ICT), Korea, under the ITRC(Information Technology Research Center) support program(IITP-2025-RS-2024-00437102) and the Global Research Support Program in the Digital Field program(IITP-2025-RS-2024-00418641) supervised by the IITP(Institute for Information \& Communications Technology Planning \& Evaluation). Hengfei Wang was supported by China Scholarship Council Grant No.202006210057
\end{acks}

\balance

\bibliographystyle{ACM-Reference-Format}
\bibliography{main}

\end{document}